\documentclass[10pt,twocolumn,letterpaper]{article}

\usepackage{cvpr}              
\usepackage{multirow}
\usepackage{xcolor} 
\usepackage{CJKutf8}
\usepackage{array}

\usepackage{diagbox}
\usepackage{makecell}

\definecolor{cvprblue}{rgb}{0.21,0.49,0.74}
\usepackage[pagebackref,breaklinks,colorlinks,allcolors=cvprblue]{hyperref}

\def\paperID{2939} 
\def\confName{CVPR}
\def\confYear{2026}

\title{EmoVerse: A Large-Scale MLLM-Powered Dataset for Explainable Visual Emotion Understanding}

\author{
Yijie Guo$^{1}$, Dexiang Hong$^{1}$, Weidong Chen$^{1*}$, Zihan She$^{1}$,\\
Cheng Ye$^{1}$, Xiaojun Chang$^{1}$, Zhendong Mao$^{1}$, Yongdong Zhang$^{1}$\\
$^{1}$University of Science and Technology of China\\
{\tt\small \{guoyijie, hongdexiang, cn211162, kyrieye\}@mail.ustc.edu.cn}\\
{\tt\small \{chenweidong, xjchang, zdmao, zhyd\}@ustc.edu.cn}
\vspace{-10pt}
}

\begin{document}
\twocolumn[{
	\renewcommand\twocolumn[1][]{#1}
	\maketitle
	\begin{center}
		\centering
		\includegraphics[width=0.95\linewidth]{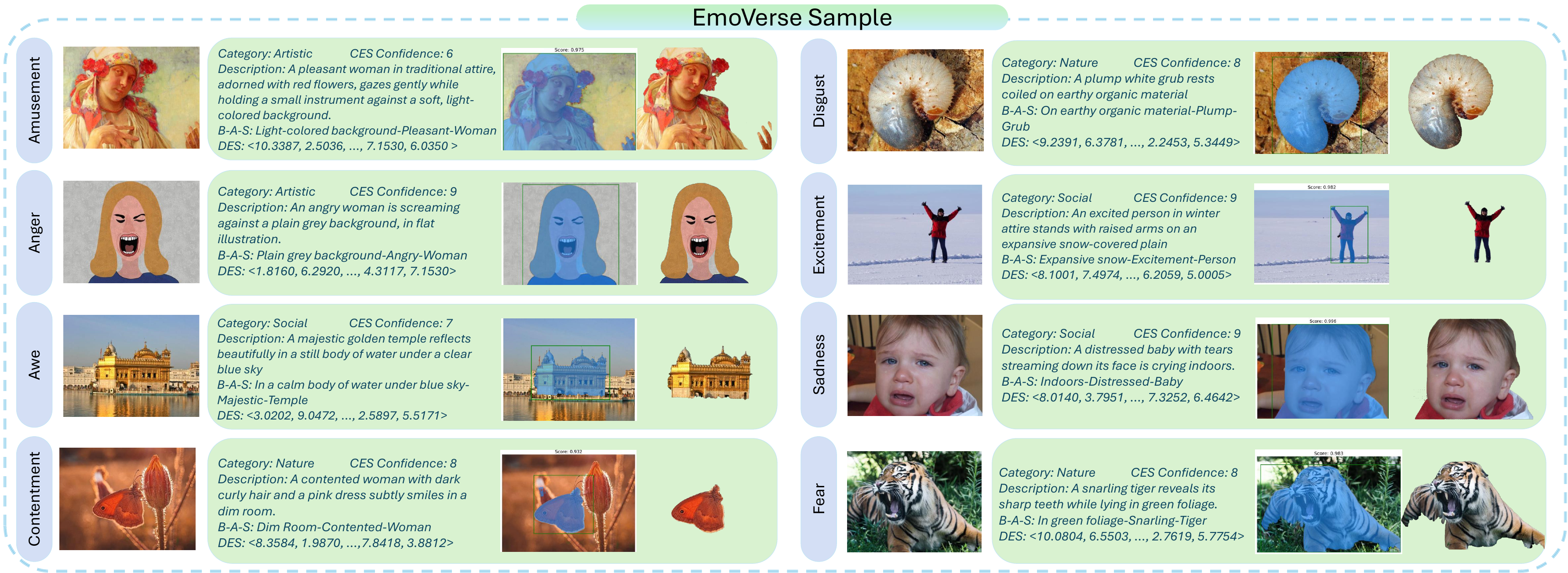}
		\captionof{figure}{EmoVerse Dataset introduces the first large-scale visual emotion dataset that combines Categorical Emotion States (CES) and Dimensional Emotion Space (DES) annotations, offering subject-level and word-level emotion attribution with various images.
		}
	\vspace{-3pt}
	\label{fig:cover}
\end{center}
}]
\begin{CJK*}{UTF8}{gbsn}
\maketitle

\renewcommand{\thefootnote}{\fnsymbol{footnote}}
\footnotetext[1]{Corresponding author}

\begin{abstract}

    \vspace{-8pt}
    Visual Emotion Analysis (VEA) aims to bridge the affective gap between visual content and human emotional responses. Despite its promise, progress in this field remains limited by the lack of open-source and interpretable datasets. Most existing studies assign a single discrete emotion label to an entire image, offering limited insight into how visual elements contribute to emotion. In this work, we introduce EmoVerse, a large-scale open-source dataset that enables interpretable visual emotion analysis through multi-layered, knowledge-graph-inspired annotations. By decomposing emotions into Background–Attribute–Subject (B-A-S) triplets and grounding each element to visual regions, EmoVerse provides word-level and subject-level emotional reasoning. With over 219k images, the dataset further includes dual annotations in Categorical Emotion States (CES) and Dimensional Emotion Space (DES), facilitating unified discrete and continuous emotion representation. A novel multi-stage pipeline ensures high annotation reliability with minimal human effort. Finally, we introduce an interpretable model that maps visual cues into DES representations and provides detailed attribution explanations. Together, the dataset, pipeline, and model form a comprehensive foundation for advancing explainable high-level emotion understanding.
\end{abstract}
    
\section{Introduction}
\label{sec:intro}

\begin{flushleft}
	\textit{``Emotions are the colors of life; without them, we would live in a gray world.''}
\end{flushleft}
\vspace{-18pt}
\begin{flushright}
	\textit{--Eli Addis}
\end{flushright}
\vspace{-5pt}

Emotions are fundamental to human intelligence, influencing cognition, perception, and interaction. A long-standing goal in Artificial Intelligence (AI) is to endow machines with the ability to perceive, understand, and respond to human emotions. With the rapid progress of Visual Language Models (VLMs)~\cite{li2022blip, radford2021learning} and Multimodal Large Language Models (MLLMs) \cite{ramesh2021zero}, Visual Emotion Analysis (VEA)~\cite{hu2025grounding, lin2025make, dang2025emoticrafter} has emerged as a key frontier that bridges visual content and affective response, reshaping the way humans engage with AI systems and multimodal agents~\cite{liu2025moee, gervasi2023applications, zhang2025emit, wang2023emotionclip}.

Despite recent advances, Visual Emotion Analysis (VEA) remains challenging due to the inherent subjectivity and complexity of human emotions~\cite{niedenthal2017psychology, ye2024dualpath, jin2024radiology}. A major reason for this challenge lies in the lack of large-scale, high-quality datasets that can accurately capture subtle and context-dependent affective cues. Existing datasets still suffer from limited scale and diversity, weak annotation reliability, and the absence of interpretable emotion grounding. As user-generated visual content and generative models proliferate~\cite{chen2025emova, hu2025music2palette, jin2024d2net}, developing a comprehensive and fine-grained understanding of emotional semantics becomes increasingly essential for high-level vision tasks such as emotion-aware editing~\cite{yang2025emoedit, wang2025emotivetalk, zhang2025creatidesign}, emotion alignment~\cite{cong2025emodubber, kim2025contextface, lin2025mvportrait, chen2026facenet}, and affect-driven visual understanding~\cite{yang2024emogen, shen2025cocoer, fang2025emoe, xia2025seek}.

To this end, we present EmoVerse dataset, a large-scale, open-source dataset designed for fine-grained and interpretable visual emotion understanding. EmoVerse deconstructs emotions into structured semantic triplets inspired by Knowledge Graphs (Background–Attribute-Subject, B-A-S) and object-level grounding via Grounding DINO~\cite{liu2024grounding} and SAM~\cite{kirillov2023segment}, linking contextual, attribute, and subject elements for interpretable affective reasoning. Each image is annotated with both Categorical Emotion States (CES)~\cite{ps2017emotion} and Dimensional Emotion Space (DES)~\cite{zhao2016predicting}, enabling unified discrete and continuous emotion representation.

The construction of such a rich dataset is enabled by a novel, multi-stage Annotation and Verification Pipeline that combines advanced VLMs, EmoViT~\cite{xie2024emovit}, and a Chain-of-Thought (CoT)–based Critic Agent~\cite{wei2022chain} to ensure annotation reliability. Finally, we fine-tuned Qwen2.5-VL~\cite{bai2025qwen2} to develop a high-dimensional emotion projector, mapping visual cues into a 1024-dimensional emotion embedding.

In summary, our contributions are:

\begin{itemize}
	\setlength{\itemsep}{0pt}
	\setlength{\parsep}{0pt}
	\setlength{\parskip}{0pt}
	
	\item We present EmoVerse, the first large-scale visual emotion dataset that offers high-dimensional DES annotations together with rich, fine-grained B-A-S triplets and object-level grounding, surpassing existing VEA datasets in scale, annotation richness, and diversity.
	
	\item We propose a novel Annotation and Verification Pipeline that ensures high-quality and consistent data annotations with minimal human intervention.
	
	\item We develop an interpretable emotion model that maps visual cues into a continuous DES space for DES representations and provides detailed, interpretable attribution explanations for advanced VEA tasks.
	
\end{itemize}

\section{Related Work}
\label{sec:rw}

\subsection{Visual Emotion Datasets}

Emotion models in psychology are generally divided into Categorical Emotion States (CES) \cite{ps2017emotion} and Dimensional Emotion Space (DES) \cite{zhao2016predicting}. CES models, such as Mikels’ eight categories \cite{mikels2005emotional}, use discrete and interpretable labels, suitable for classification but limited in expressing mixed or subtle emotions. DES models, by contrast, represent emotions as points in a continuous space, providing richer affective granularity for regression-based analysis~\cite{ye2025emotioncause}.

Early datasets like Flickr and Instagram \cite{katsurai2016image} collected web images using emotion keywords and binary sentiment labels. FI dataset \cite{you2016building} extended this to 23k labeled samples with eight categories. Subsequent works, such as EmoSet \cite{yang2023emoset} and EmoArt \cite{zhang2025emoart}, enlarged scale and diversity by combining human and MLLM annotations~\cite{liu2025street2sat}, introducing auxiliary attributes like scene type to improve interpretability~\cite{zhou2025distillation, ye2025videosum}.

\begin{table*}[t]
    \setlength{\abovecaptionskip}{0.1cm}
    \setlength{\belowcaptionskip}{0.1cm}
    \scriptsize
    \renewcommand\arraystretch{1.2}  
    \setlength\tabcolsep{4pt}       
    \centering
    \caption{Comparison of emotion-related datasets and their annotation characteristics.}
    \label{tab:dataset_comparison}
    \begin{tabular}{lcccccccccc}
        \toprule[1pt]
        Dataset & \#Image & Label Source & Tasks & Image Type & Category & Description & Word-level Anno. & Category Conf. & Subject-level Anno. \\
        \midrule[0.5pt]
        FI~\cite{you2016building} & 23K & Human & R & Social & CES(Sentiment-2) & $\times$ & $\times$ & $\times$ & $\times$ \\
        Instagram~\cite{katsurai2016image} & 42K & Human & R & Social & CES(Sentiment-2) & $\times$ & $\times$ & $\times$ & $\times$ \\     
        Emotion6~\cite{peng2015mixed} & 1.98K & Human & R & Social & CES(Ekman-6) & $\times$ & $\times$ & $\times$ & $\times$ \\
        FindingEmo~\cite{mertens2024findingemo} & 25K & Human & R & Social & CES(Plutchik-8) & $\times$ & $\times$ & $\times$ & $\times$ \\
        Artemis~\cite{achlioptas2021artemis} & 80K & Human & G\&R & Artistic & CES(Mikels'-8) & $\checkmark$ & $\times$ & $\times$ & $\times$ \\
        EmoSet~\cite{yang2023emoset} & 118K & Human\&LLM & G\&R & Social\&Artistic & CES(Mikels'-8) & $\times$ & $\times$ & $\times$ & $\times$ \\
        EmoArt~\cite{zhang2025emoart} & 130K & Human\&LLM & G\&R & Artistic & CES(12) & $\checkmark$ & $\times$ & $\times$ & $\times$ \\
        \textbf{EmoVerse (Ours)} & \textbf{219K} & \textbf{Human\&LLM} & \textbf{G\&R} & \textbf{Social\&Artistic} & \textbf{CES(Mikels'-8)\&DES} & \textbf{$\checkmark$} & \textbf{$\checkmark$} & \textbf{$\checkmark$} & \textbf{$\checkmark$} \\
        \bottomrule[1pt]
    \end{tabular}
    \vspace{-10pt}
\end{table*}

Despite progress, existing works still face key issues: limited scale and diversity, weak affective reliability, and absence of fine-grained cues or subject-level grounding. Most provide only discrete labels without contextual or intensity information, making it difficult to model nuanced emotions. In light of this, we construct EmoVerse dataset to bridge the gap, the comparison is provided in \Cref{tab:dataset_comparison}.

\subsection{Dataset Annotation and Verification}

The construction of high-quality datasets has been widely recognized as a crucial foundation for advancing research in computer vision and affective computing. Early efforts in dataset development often relied on manual annotation without systematic verification~\cite{russell2008labelme}, which raised concerns about annotation noise and label consistency. To address this, crowd-sourced labeling platforms, such as Amazon Mechanical Turk~\cite{crowston2012amazon}, have been widely adopted, enabling large-scale data collection with reduced cost and time. Several studies have further emphasized the importance of annotation reliability by introducing strategies such as majority voting, label aggregation, and inter-rater agreement metrics to mitigate subjectivity and ensure robustness~\cite{snow2008cheap, deng2009imagenet, raykar2010learning}. More recent works have explored semi-automatic annotation pipelines, leveraging pre-trained models and strategies to minimize labeling errors~\cite{bearman2016s, liu2024radiologyllm}.

Alongside annotation, verification procedures have increasingly focused on quality control mechanisms, including redundancy in labeling, expert verification~\cite{welinder2010online, huang2025gmoe}, and cross-verification~\cite{krishna2017visual, song2025retrieval}. Collectively, these approaches demonstrate a clear trend toward balancing scalability with reliability in dataset construction, underscoring the need for well-defined annotation and verification workflows~\cite{russakovsky2015imagenet, li2025cir, lin2024multihop}.

Building upon these insights, an automated annotation and verification pipeline emerges as a promising direction for achieving large-scale, high-fidelity dataset construction—enabling scalable annotations while maintaining data accuracy and reducing manual effort.

\subsection{Emotion Representation}

Recent advances in Vision–Language Models (VLMs) have demonstrated that large-scale multimodal pre-training can endow models with impressive visual–semantic reasoning abilities~\cite{yin2025knowledge, yang2025uncertain, wu2025emotion, zhu2025emosym}. However, the latent spaces of most VLMs are primarily optimized for generic alignment tasks such as image captioning or question answering, rather than for capturing the emotional semantics embedded in visual content, which compresses image features into text-aligned embeddings without explicitly modeling emotional intensity~\cite{chen2026soecgn}, category relations, or subject grounding~\cite{jia2021scaling, zhang2024open}.

On the other hand, emotion representation learning aims to encode affective information within a continuous space. Early image datasets, like Flickr30k~\cite{katsurai2016image} and FindingEmo~\cite{mertens2024findingemo}, primarily focused on descriptive annotations that capture object-level or scene-level semantics rather than affective cues~\cite{fu2024livecomment, wang2023contour}. Subsequent works began to incorporate emotion-related attributes, with datasets like EmoSet introducing auxiliary annotations such as image brightness, colorfulness, and human actions to approximate emotional content~\cite{yang2023emoset, li2024graphcaption}. However, some of these annotations often fail to capture the complex, background-dependent nature of human emotions, limiting their effectiveness in detailed visual emotion analysis.

To bridge this gap, we develop an emotion model that maps visual cues into interpretable affective representations, providing high-dimensional DES representations and detailed emotion attribution explanations.

\section{Methods}
\label{sec:methods}

\subsection{EmoVerse Dataset}

\begin{figure*}
	\centering
	\includegraphics[width=\linewidth]{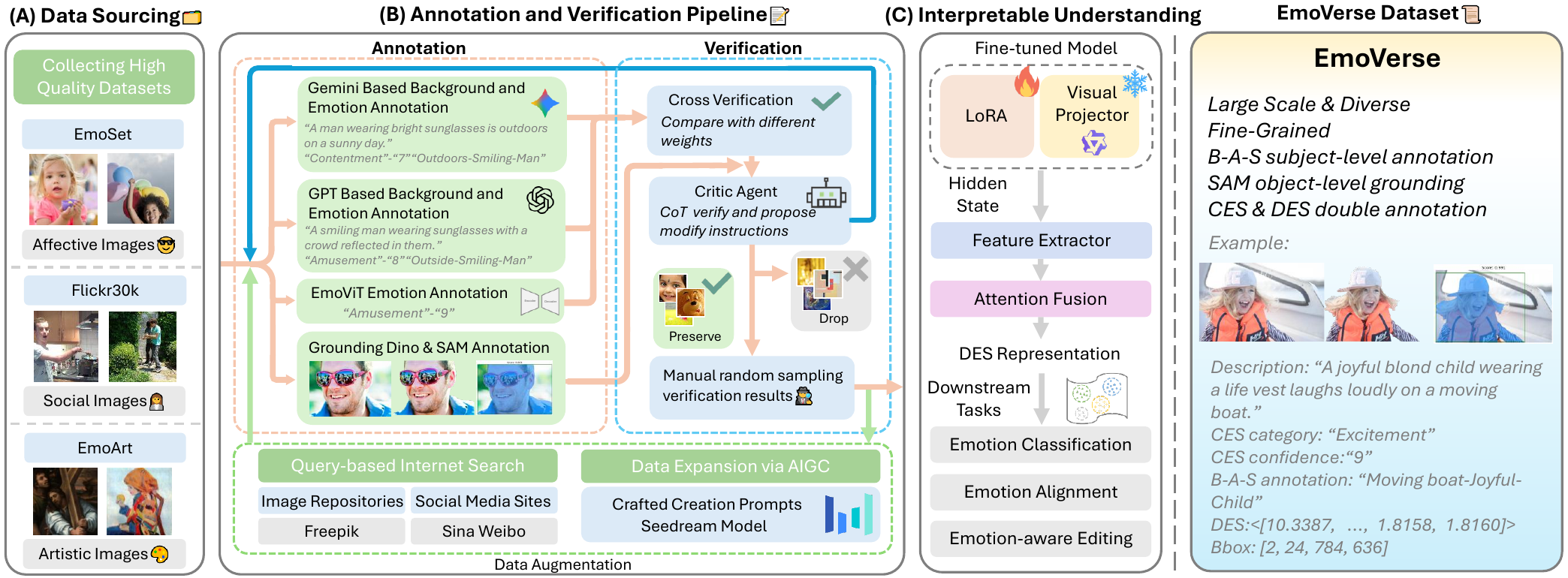}
	\caption{Overview of EmoVerse. EmoVerse collects images from multiple sources. Images collected pass through Annotation and Verification Pipeline. DES annotations are generated from our Interpretable Model, enabling unified understanding of visual emotions.
	}
	\label{fig:overall}
    \vspace{-5pt}
\end{figure*}

The EmoVerse dataset involved two core stages: a hybrid data sourcing and integration strategy to ensure scale and diversity, and the implementation of a novel, multi-layered annotation schema to capture fine-grained emotional cues, the process is shown in ~\Cref{fig:overall}.

\subsubsection{Data Sourcing and Integration}

Unlike datasets constructed solely through keyword-based web search queries, EmoVerse consists of three parts: images from existing datasets, images collected from the Internet, and AIGC expansions.

\smallskip

\noindent\textbf{Integration and Refining Existing Datasets.} We architected the dataset through the strategic integration and filtration of several high-quality, large-scale public datasets. Each source is chosen for its unique contribution to the final dataset's breadth and depth:

\begin{itemize}
	\setlength{\itemsep}{0pt}
	\setlength{\parsep}{0pt}
	\setlength{\parskip}{0pt}
	
	\item \textbf{EmoSet}~\cite{yang2023emoset}. This dataset acts as the emotional foundation of EmoVerse. As a large-scale visual emotion dataset with carefully verified human annotations, it offers a dependable base of images with confirmed emotional labels based on Mikels' eight category model.
	
	\item \textbf{EmoArt}~\cite{zhang2025emoart}. To ensure stylistic diversity and prevent our models from overfitting on photorealistic images, we integrated artistic datasets, compelling models to learn emotional cues from fundamental artistic principles such as color palettes, brushstroke textures, and abstract forms.
	
	\item \textbf{Flickr30k}~\cite{young2014image}. Flickr30k offers a rich collection of natural, real-world images with descriptive captions, which are crucial for learning visual-semantic alignments.
	
\end{itemize}

\smallskip

\noindent\textbf{Augmentation via Web-Sourced Imagery.} To ensure coverage of long-tail concepts and contemporary visual trends, the second part of EmoVerse consists of images collected from the Internet. Our collection strategy was designed to be more targeted than traditional broad keyword searches.

\textbf{(1) Query Generation:}
We leveraged our B-A-S semantic triplets to generate highly specific search queries such as "joyful crowd at music festival". Query phrases are derived from processed images that have been sorted through our pipeline. This method yields images with much higher relevance to specific, nuanced emotional contexts.

\textbf{(2) Image Collection:}
Images were retrieved from multiple online platforms, including royalty-free stock image repositories (such as Freepik\footnotemark[1]) and social media sites, to capture a wide variety of subjects, compositions, and photographic styles. We also verify the images using open-source models on GitHub after collection to ensure they are not duplicates of existing images. 

\smallskip

\noindent\textbf{Dataset Enrichment
via AIGC.} To further enhance dataset diversity and demonstrate the extensibility of our B-A-S (Background-Attribute-Subject) framework, we introduced a third data source: AI-Generated Content. We leveraged our annotated B-A-S triplets as seed prompts. By systematically replacing one or two elements within these triplets, we generated new, targeted compositional prompts. Using the Seedream model~\cite{seedream2025seedream}, we synthesized approximately 25,000 images from these prompts. This AIGC subset, accounting for 12.17\% of our total dataset, significantly enriches the coverage of emotional concepts and effectively populates long-tail emotional scenarios that are difficult to capture or rarely found in real-world images.

In conclusion, our data collection strategy provides two primary advantages: diversity and quality. By purposefully merging varied sources—from the artistic works in EmoArt to the naturalistic images in Flickr30k and affective images in EmoSet, we have built a visually heterogeneous dataset that mitigates stylistic overfitting. This diversity is further enhanced by our targeted, B-A-S-based web search and AIGC enrichment, which captures specific, long-tail emotional concepts and enriches concept coverage.

\footnotetext[1]{\url{https://www.freepik.com/}}

\subsubsection{Fine-Grained Annotation and Multi-Dimensional Representation}

EmoVerse dataset provides multi-stage annotations, designed to bridge the affective gap between low-level pixels and high-level human emotion in an interpretable way.

\smallskip

\noindent\textbf{Knowledge-Graph-Inspired Semantic Annotation.} The Background–Attribute–Subject (B-A-S) triplet serves as a minimal emotional knowledge unit, decomposes an image’s emotional content into semantic components.

This decoupled structure provides word-level supervision, explicitly grounding contextual, attribute, and subject cues to distinct visual regions. Such alignment enhances the model’s understanding of how individual elements collectively shape emotion. Elements can also be recombined to synthesize new emotions, providing high flexibility.

\smallskip

\noindent\textbf{CES and DES Annotations.} Moving beyond the limitations of discrete emotion categories, EmoVerse provides a continuous, multi-dimensional representation of affect. 

For Categorical Emotion Space (CES)~\cite{ps2017emotion}, we adopt Mikels’ eight-class model (amusement, awe, contentment, excitement, anger, disgust, fear, and sadness) and provide confidence scores indicating the clarity of each emotion.

Complementing CES, the Dimensional Emotion Space (DES)~\cite{zhao2016predicting} projects each image into a 1024-dimensional embedding using our Interpretable Model. This enables fine-grained emotion intensity estimation, smooth interpolation between emotions, and quantitative measurement of affective distance between images. DES further enhances downstream emotion understanding by fostering richer, more robust, and generalizable feature learning.

\smallskip

\noindent\textbf{Subject-level Instance Annotation.} To semantically ground our B-A-S labels directly to image regions, we employed Grounding DINO with the Segment Anything Model (SAM). For every image, the primary subject identified in the annotation is precisely localized with bounding boxes and segmentation masks. This links the abstract textual labels and emotion scores to the specific group of pixels that represent the subject, enabling models to learn which object, in what state, evokes a particular emotion.

\subsection{Cross Verification Pipeline}
\label{sec:Annotation and Verification}

To ensure the high quality and accuracy of our dataset, we implemented a multi-stage pipeline for data annotation and verification, the process is shown in ~\Cref{fig:overall} part B. This process leverages multiple advanced AI models for initial annotation, followed by a verification protocol involving a Critic Agent and human oversight.

We first employed two state-of-the-art Visual Language Models, Gemini 2.5 and GPT-4o, to annotate background context and emotional sentiment and make comparisons. Since LLMs are not entirely accurate for sentiment understanding~\cite{bhattacharyya2025evaluating}, the comparison results of emotional labels and emotion confidence scores are compared against the outputs from EmoViT, which has been previously verified to be more accurate in sentiment labeling~\cite{xie2024emovit}, thus carrying greater weight in comparison.

To further enhance annotation reliability, we introduce a Critic Agent that acts as an independent quality inspector within the verification loop. The Critic Agent uses a Chain-of-Thought (CoT) \cite{wei2022chain} reasoning framework that decomposes verification into a series of clear analytical steps. For each sample, the agent first analyzes the rough scene description. Then, it progressively examines its consistency with the background caption and emotion label through explicit reasoning steps. Based on the inferred reasoning chain, each annotation is classified as valid, revisable, or discarded. When revisions are required, the Critic Agent produces modification instructions that are then fed back into the annotation module during the next iteration. However, due to the subjectivity of emotion intensity, the Critic Agent only supervises emotion intensity at three discrete levels: high, medium, and low, without evaluating its exact numerical value. This process allows the pipeline to maintain high semantic fidelity and contextual coherence with minimal human intervention, providing a crucial foundation for the reliability of the EmoVerse dataset. Finally, a subset of samples underwent human inspection as a ground-truth check, ensuring alignment with human judgment and providing a quantitative measure of dataset reliability.

\subsection{Interpretable Model}
\label{sec:Prjector}
\begin{figure}
	\centering
	\includegraphics[width=0.97\linewidth]{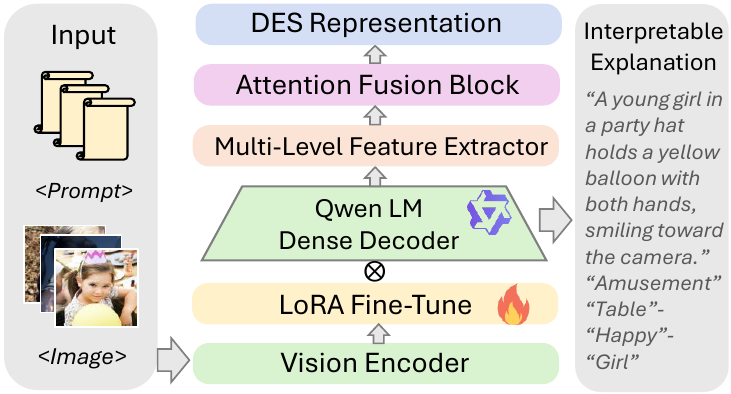}
	\caption{Architecture of our Interpretable Model. Model fine-tunes Qwen model to acquire explanation and incorporates Feature Extractor and Attention Block to acquire DES representation.
	}
	\label{fig:model}
    \vspace{-15pt}
\end{figure}

To enable interpretable understanding, we introduce a two-stage training framework based on Qwen2.5-VL-3B~\cite{bai2025qwen2}. The overall process is illustrated in \Cref{fig:model}. The projector was first fine-tuned through a two-round training process to improve both emotional attribution and categorical accuracy. In the first round, the model is fine-tuned using the attribute annotations from our dataset. In the second round, we further fine-tuned the model with emotion category labels to better understand high-level emotion meanings and improve overall classification stability. Throughout the training process, the model receives images $I$ and prompts $P$ as inputs and is trained to output explanations. The model is optimized using cross-entropy loss, enabling it to learn how visual cues contribute to emotional perception.
\begin{equation}
    \mathcal{L}_{\text{CE}} = - \sum_{t} y_t \log \hat{y}_t,
    \label{eq:ce}
\end{equation}
After fine-tuning, the trained model acts as a frozen feature interpreter, with the generated embeddings first passing through the feature extractor, where the last four transformer layers of Qwen2.5-VL are extracted, then through pooling and projection layer.
\begin{equation}
\mathbf{f}_{\text{proj}}
    = \mathbf{W}_2\,\phi\!\Big(
        \mathbf{W}_1 
        \Big(\sum_{k=0}^{3}\alpha_k\,\bar{\mathbf{h}}_{L-k}\Big)
        + \mathbf{b}_1
      \Big) + \mathbf{b}_2,
\label{eq:fusion_projection}
\end{equation}

where $\mathbf{h}_i$ denotes the hidden representation from the $i$-th transformer layer, $L$ is the final transformer layer. $\mathbf{W}_1 \in \mathbb{R}^{H \times H}$ and $\mathbf{W}_2 \in \mathbb{R}^{H \times \frac{H}{2}}$ are learnable parameters that reduce the dimensionality from $H$ to $H/2$ while preserving expressive capacity. $\boldsymbol{\alpha}$ are learnable weights that adaptively aggregate the last four layers. $\phi(\cdot)$ denotes a nonlinear activation function, and Dropout is applied between the two projection layers for regularization. This aggregation captures both high-level semantics and intermediate perceptual cues essential for emotion interpretation.

After extraction, we employ an attention-based fusion block that performs feature fusion, adaptively weighting sequence elements according to their emotional relevance. The attended outputs are then pooled through weighted averaging to produce the DES representation.
\begin{align}
	\label{eq:a_self}
	\mathbf{A}_{s} 
	&= \textit{softmax}\!\left(
	\frac{(\mathbf{f}_{\text{proj}} W_{q}^{s}) 
	(\mathbf{f}_{\text{proj}} W_{k}^{s})^{\!T}}
	{\sqrt{d_k}}
	\right)
	(\mathbf{f}_{\text{proj}} W_{v}^{s}), \\
	\label{eq:a_cross}
	\mathbf{A}_{c} 
	&= \textit{softmax}\!\left(
	\frac{(\mathbf{A}_{s} W_{q}^{c})
	(\mathbf{f}_{\text{proj}} W_{k}^{c})^{\!T}}
	{\sqrt{d_k}}
	\right)
	(\mathbf{f}_{\text{proj}} W_{v}^{c}).
\end{align}
where $W_{q}^{s}$, $W_{k}^{s}$, $W_{v}^{s}$ are the learned projection matrices in the self-attention block, $W_{q}^{c}$, $W_{k}^{c}$, $W_{v}^{c}$ are parameters in the cross-attention block, and $d_k$ denotes the key dimension for normalization. This produces our DES representation, providing a continuous and interpretable representation of visual emotions for downstream tasks such as emotion classification, retrieval, and generation.

\begin{figure}
	\centering
	\includegraphics[width=0.95\linewidth]{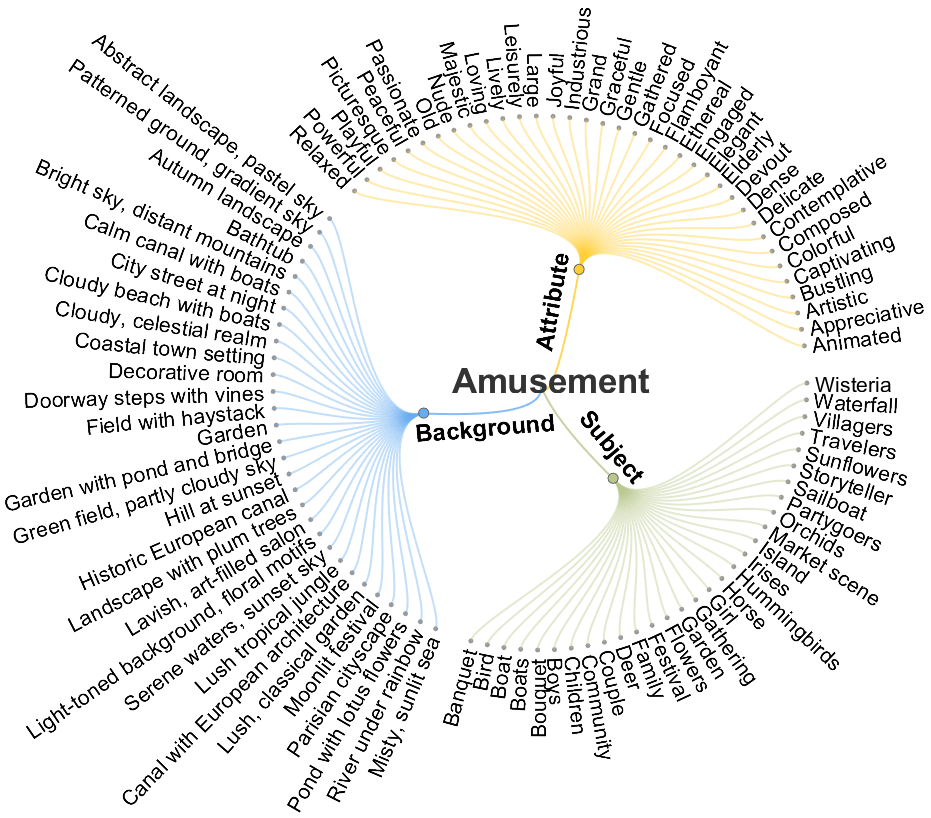}
	\caption{Knowledge graph based on B-A-S triplets. Decoupled representation facilitates emotion attribution and provides extensibility for understanding and generating diverse affective scenarios.
	}
	\label{fig:knowledge}
\end{figure}

\begin{table}[t]
    \centering
    \scriptsize
    \caption{Cross-dataset generalization results, reported in Top-1 accuracy (\%). EmoVerse-Cut is a random subset at EmoSet size.}
    \label{tab:cross_verification}
    \renewcommand\arraystretch{1.25}

    \begin{tabular}{
        >{\raggedright\arraybackslash}p{1.5cm}|
        >{\centering\arraybackslash}p{1cm}
        >{\centering\arraybackslash}p{1cm}
        >{\centering\arraybackslash}p{1cm}
        >{\centering\arraybackslash}p{1cm}
    }
        \specialrule{0.75pt}{1pt}{1.2pt}
        \multicolumn{5}{c}{\textbf{w/o pretrained}} \\
        \specialrule{0.5pt}{1pt}{1pt}
        \diagbox[dir=NW]{Test}{Train} & FI & EmoSet & \makebox[0pt][c]{EmoVerse-Cut} & EmoVerse \\
        \Xhline{0.5pt}
        FI & 40.62 & 35.26 & 36.09 & \textbf{43.03} \\
        EmoSet & 24.50 & 51.48 & 56.58 & \textbf{73.50} \\
        EmoVerse & 35.02 & 41.35 & 49.32 & \textbf{68.97} \\
        \specialrule{0.75pt}{1.2pt}{1pt} 
    \end{tabular}

    \begin{tabular}{
        >{\raggedright\arraybackslash}p{1.5cm}|
        >{\centering\arraybackslash}p{1cm}
        >{\centering\arraybackslash}p{1cm}
        >{\centering\arraybackslash}p{1cm}
        >{\centering\arraybackslash}p{1cm}
    }
        \multicolumn{5}{c}{\textbf{w/ pretrained (ImageNet)}} \\
        \specialrule{0.5pt}{1pt}{1pt}
        \diagbox[dir=NW]{Test}{Train} & FI & EmoSet & \makebox[0pt][c]{EmoVerse-Cut} & EmoVerse \\
        \Xhline{0.5pt}
        FI & \textbf{66.25} & 48.44 & 51.64 & 56.57 \\
        EmoSet & 47.81 & 72.92 & 79.24 & \textbf{81.87} \\
        EmoVerse & 37.23 & 63.05 & 69.51 & \textbf{72.14} \\
        \specialrule{0.75pt}{1.2pt}{1pt}
    \end{tabular}

    \vspace{-6pt}
\end{table}

\begin{table*}[t]
    \centering
    \scriptsize
    \caption{Evaluation of the Annotation and Verification Pipeline. Verified Data part reports the component ablation results and Critic Agent ability on the human-verified subset. Corrupted Data part reports Critic Agent’s recall rate on deliberately corrupted annotations.}
    \label{tab:ablation_study}
    \renewcommand\arraystretch{1.18}

    \begin{tabular}{l|cccc|c}
        \toprule
        \multirow{2}{*}{\textbf{Attribute}} 
        & \multicolumn{4}{c|}{\textbf{Verified Data}} 
        & \textbf{Corrupted Data} \\
        \cline{2-6}
        & Full Pipeline Acc. 
        & w/o Cross-Verif. Acc. 
        & w/o Critic Agent Acc. 
        & Critic Agent Preserve Rate 
        & Critic Agent Recall Rate \\ 
        \midrule

        Emotion Category  & \textbf{93.20} & 80.53 & 72.50 & 99.72 & 89.65 \\
        Description       & \textbf{90.56} & 83.11 & 69.93 & 96.12 & 97.27 \\
        B-A-S Triplet     & \textbf{96.16} & 85.40 & 60.32 & 90.50 & 85.78 \\
        Emotion Intensity & \textbf{71.14} & 70.21 & 65.83 & 85.61 & 45.79 \\
        Bounding Box      & \textbf{85.46} & --    & 75.77 & 86.38 & 78.42 \\

        \bottomrule
    \end{tabular}

    \vspace{-5pt}
\end{table*}

\section{Analysis of EmoVerse}
\label{sec:experiment}

\subsection{Evaluation of EmoVerse Dataset}

\subsubsection{Datasets Comparison}

EmoVerse seeks to build a comprehensive and interpretable dataset to assist researchers. To the best of our knowledge, this is the first large-scale VEA dataset annotated in both CES and DES. EmoVerse offers advantages over existing datasets in four key areas: scale, diversity, unique annotations, and annotation accuracy.

\noindent\textbf{Scale.} Our dataset comprises over 218,522 finely annotated images, significantly surpassing the scale of previously existing large-scale datasets. Specifically, it is approximately 1.9 times larger than EmoSet (118,102 images) and exceeds the FI dataset (23,308 images) by over 9.4 times. This constitutes the largest visual emotion dataset with annotations in terms of total number, thus offering an unparalleled resource for training and evaluating visual emotion models.

\noindent\textbf{Diversity.} EmoVerse achieves its diversity through three approaches: (1) filtering varied, large-scale public datasets, including EmoSet, EmoArt, and Flickr30k; (2) supplementing them with new web content using crafted B-A-S-based queries. (3) controlled dataset expansion via AIGC. This combined strategy ensures our dataset includes images from various styles, such as art, nature, social media, and synthetic content, helping prevent style overfitting. In~\Cref{fig:knowledge}, we present the knowledge graph of our B-A-S triplets, highlighting the diversity of our dataset.

\noindent\textbf{Unique Annotation.} EmoVerse offers a multi-layered annotation schema far richer than existing datasets. A key innovation is the Background-Attribute-Subject (B-A-S) triplet, which deconstructs emotion into its semantic components. Unlike single-label datasets, EmoVerse is the first to provide annotations for both discrete Categorical Emotion States (CES) and Dimensional Emotion Space (DES). We also include a confidence score for emotion categories and ground its semantic labels at the subject level using Grounding DINO and Segment Anything Model to produce accurate bounding boxes and segmentation masks.

\begin{figure*}
	\centering
	\includegraphics[width=\linewidth]{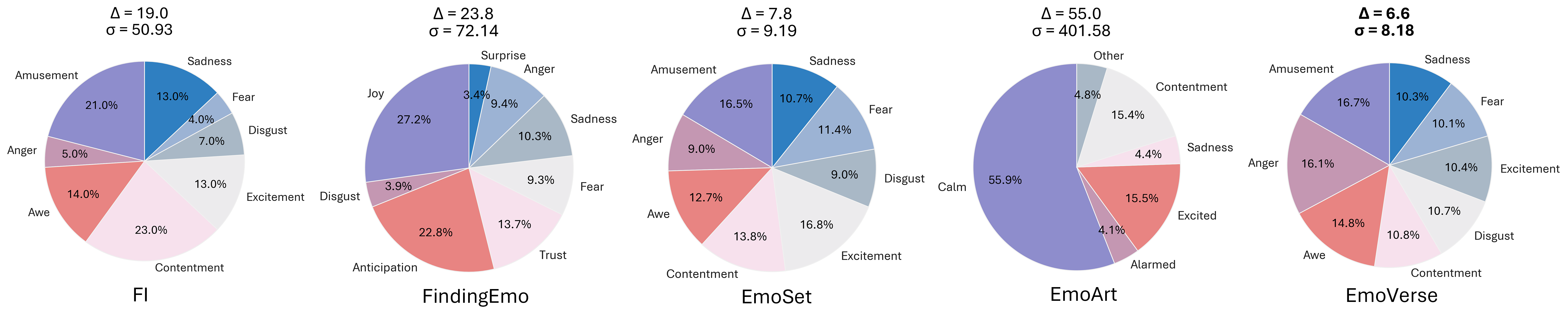}
	\caption{Emotion category distribution statistics. Colored segments show the percentage of each category. $\Delta$ is the minimum and maximum difference and $\sigma$ is the variance. EmoVerse dataset shows great balance in emotion distribution.
	}
	\label{fig:distribution}
    \vspace{-10pt}
\end{figure*}

\noindent\textbf{Annotation Accuracy and Balance.} By integrating a multi-stage Annotation and Verification Pipeline, EmoVerse achieves high annotation accuracy and consistency. Both the ablation study (\Cref{tab:ablation_study}) and user study (\Cref{tab:user_study}) validate the reliability of our annotations, outperforming existing VEA datasets. Moreover, EmoVerse mitigates the critical issue of data imbalance found in prior works, as shown in \Cref{fig:distribution}, our dataset exhibits the minimum of maximum difference ($\Delta$) and variance ($\sigma$) across categories, which is essential for training unbiased emotion recognition models.

\begin{table}[t]
	\centering
	\scriptsize
	\caption{User study results comparing annotation accuracy and consistency across datasets. EmoVerse achieves the highest scores in emotion arousal and labeling reliability.}
	\label{tab:user_study}
	\renewcommand\arraystretch{1.18}
	\begin{tabular}{lccc}
		\toprule
		Dataset & Emotion Arousal & CES Acc. & Anno. Acc. \\
		\midrule
		Flickr~\cite{young2014image} & 72.67 & 66.25 & - \\
        Emotion6~\cite{peng2015mixed} & 72.17 & 68.17 & - \\
		EmoSet~\cite{yang2023emoset} & 78.08 & 76.00 & - \\
		EmoArt~\cite{zhang2025emoart} & 64.17 & 64.25 & 82.17 \\
		EmoVerse & \textbf{82.41} & \textbf{81.83} & \textbf{86.41} \\
		\bottomrule
	\end{tabular}
\vspace{-5pt}
\end{table}

\begin{table}[t]
    \centering
    \scriptsize
    \caption{Quantitative evaluation of the Interpretable Model before and after fine-tuning on the EmoVerse dataset.}
    \renewcommand\arraystretch{1.18}
    \begin{tabular}{lccl}  
    \toprule
    \textbf{Metric} & Qwen2.5-vl & Fine-tuned & \multicolumn{1}{c}{$\Delta$} \\  
    \midrule
    BBox IoU $\uparrow$ & 74.87 & \textbf{79.24} & $\uparrow$ 4.37 \\
    BBox Center Dist $\uparrow$ & 93.06 & \textbf{94.31} & $\uparrow$ 1.25 \\
    F1 $\uparrow$ & 80.33 & \textbf{84.60} & $\uparrow$ 4.27 \\
    CLIP Score $\uparrow$ & 83.27 & \textbf{93.94} & $\uparrow$ 10.67 \\
    Emotion Acc $\uparrow$ & 41.20 & \textbf{73.43} & $\uparrow$ 32.23 \\
    Intensity Acc $\uparrow$ & 86.12 & \textbf{91.20} & $\uparrow$ 5.08 \\
    \bottomrule
    \end{tabular}
    \label{tab:model_eval}
    \vspace{-5pt}
\end{table}

\begin{figure*}
	\centering
	\includegraphics[width=0.95\linewidth]{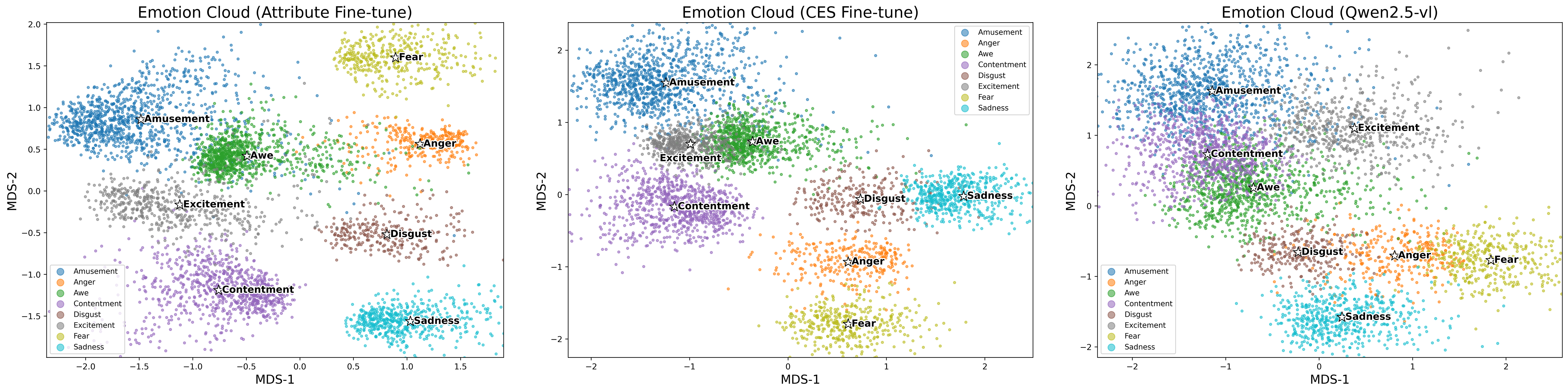}
	\caption{Visualization of emotion cloud use DES embeddings, projected through MDS. DES trained with full attribution exhibits the most compact and clearly separable clusters, reflecting that attribute guidance effectively enhances the interpretability and structural organization.
	}
	\label{fig:cloud}
    \vspace{-10pt}
\end{figure*}

\subsubsection{Cross-dataset Generalization}

To further evaluate the generalization ability and emotional distinctiveness of different visual emotion datasets, we perform a systematic cross-dataset classification experiment. Specifically, we train a ResNet-50 backbone on each dataset and evaluate its recognition accuracy on other datasets under two configurations: (1) Without Pretraining, where the model is trained from randomly initialized weights; (2) With Pretraining, where the model is initialized with ImageNet-pretrained weights. We perform training and evaluation on three representative datasets: FI~\cite{you2016building}, EmoSet~\cite{yang2023emoset}, EmoVerse-Cut (\textit{a random subset of EmoVerse dataset matched in size with EmoSet}), and EmoVerse. Each evaluation dataset is randomly sampled to contain 10,000 images from its original data distribution. We repeat the testing process five times with different random seeds and report the average Top-1 accuracy to reduce sampling bias.

As shown in \Cref{tab:cross_verification}, the results show models trained on the EmoVerse dataset achieve the highest cross-dataset recognition accuracy in both settings, demonstrating that the EmoVerse dataset offers emotionally salient and transferable representations. Even when its size is controlled (EmoVerse-Cut), performance remains consistently strong, confirming that the observed improvements come from annotation and attribution quality rather than dataset size.

\subsection{Evaluation of Verification Pipeline}

\subsubsection{Component Evaluation}

To further evaluate the robustness of our Annotation and Verification Pipeline, we designed complementary experiments focusing on data reliability and system ablation. We performed two complementary analyses: (1) A component ablation on the human-verified subset to assess how removing Cross-Verification or the Critic Agent affects the preservation of correct annotations. (2) An error-recall evaluation on a deliberately corrupted dataset. The result is shown in~\Cref{tab:ablation_study}. Bounding Box is annotated by Grounding Dino, thus it doesn't pass Cross-Verification module. 

\begin{itemize}
	\setlength{\itemsep}{0pt}
	\setlength{\parsep}{0pt}
	\setlength{\parskip}{0pt}
	
	\item Component Ablation: Ablation study reports results on the human-verified dataset when removing either Cross-Verification or the Critic Agent. The results demonstrate that (i) Cross-Verification effectively reduces inter-model bias, and (ii) the Critic Agent, with its CoT verification, is essential for maintaining semantic consistency, especially for background-related attributes.
	
	\item Error Recall on Corrupted Data: We further evaluate the Critic Agent’s capability to detect incorrect annotations. The agent shows strong recall for semantic and contextual errors, proving the capability of our Critic Agent and pipeline. The recall rate for Corrupted Emotion Intensity is relatively lower, primarily because emotion intensity is inherently subjective and difficult to evaluate consistently across samples. Therefore, in our pipeline, the Critic Agent only verifies emotion intensity in discrete levels, instead of predicting exact numerical values. The high recall rate achieved on other quantitative attributes demonstrates the robustness and reliability of our Annotation and Verification Pipeline.
\end{itemize}

\subsubsection{User Study}

To further validate the reliability of our dataset annotations and the effectiveness of our proposed pipeline, we conducted a user study involving 50 participants from diverse academic backgrounds. Specifically, we designed five groups of images, each from a different dataset, with each group containing 50 randomly selected images along with their original emotion labels and background annotations. Participants were asked to answer the following questions: (1) Can this image evoke your emotion? (2) Is the sentiment labeling of this image accurate? (3) Is the background annotation of this image accurate? Since Flickr, Emotion6, and EmoSet do not include contextual annotations, their annotation acccuracy are not reported. The results in~\Cref{tab:user_study} show that EmoVerse is the most preferred choice for all questions, confirming the accuracy of our dataset and the effectiveness of the Verification and Annotation Pipeline.

\subsection{Evaluation of Interpretable Model}

\subsubsection{Model Comparison}

To evaluate the contribution of EmoVerse attribution and the effectiveness of our fine-tuned model, we performed model comparisons between Qwen2.5-VL and our emotion-enhanced model trained on the EmoVerse dataset. The training goal is to improve the model’s ability to understand and attribute emotions across visual scenes. The evaluation metrics include Bbox IoU, Center Distance, and PRF for assessing spatial grounding accuracy, CLIP scores for measuring visual-textual semantic alignment, and Emotion Score and Intensity for assessing affective understanding and attribution, as shown in ~\Cref{tab:model_eval}. Specifically, the fine-tuned model exhibits substantial improvements in grounding accuracy (IoU +4.4\%), semantic alignment (CLIP score +6.4\%), and emotion understanding (Emotion Score +32.2\%). These gains demonstrate that the enriched, well-balanced annotations in EmoVerse provide more discriminative supervision signals, enabling the model to associate visual details with affective semantics better. Moreover, improvements in intensity estimation and center distance suggest that the model not only recognizes emotional categories more accurately but also learns to localize emotional cues within scenes more precisely.

\begin{table}[t]
    \centering
    \scriptsize
    \caption{Comparison of model performance under different training settings. The attribute-based fine-tuning achieves the best overall results in accuracy and consistency.}
    \label{tab:model_com}
    \renewcommand\arraystretch{1.2}
    \begin{tabular}{lcccc}
        \toprule
        \textbf{Training Setting} & \textbf{Acc. (\%)} & \textbf{Precision (\%)} & \textbf{Recall (\%)} & \textbf{F1 (\%)} \\
        \midrule
        Qwen2.5-vl & 55.35 & 62.64 & 56.29 & 58.26 \\
        CES Fine-tuned & 67.37 & 72.20 & 69.80 & 70.72 \\
        Attribute Fine-tuned & \textbf{73.74} & \textbf{77.86} & \textbf{75.74} & \textbf{76.21} \\ 
        \bottomrule
    \end{tabular}
    \vspace{-12pt}
\end{table}

\begin{figure}
	\centering
	\includegraphics[width=0.95\linewidth]{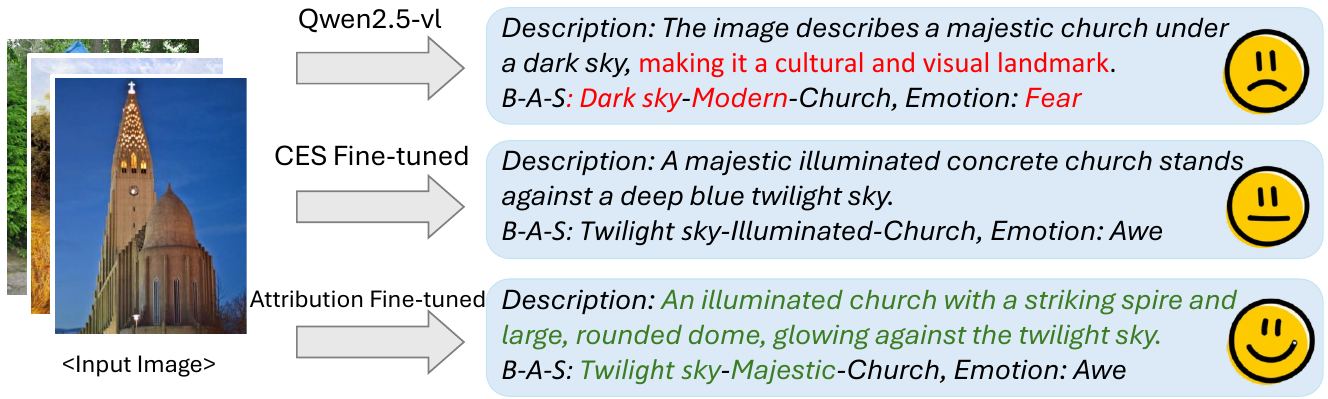}
	\caption{Visualization of comparison. The model fine-tuned by attribution shows the most accurate result.
	}
	\label{fig:illustrate}
    \vspace{-10pt}
\end{figure}

\subsubsection{Interpretability Comparison}
To further validate the interpretability and effectiveness of our DES representation, we conducted a classification experiment by attaching a linear classification head to the frozen DES embeddings for emotion category prediction. We compared our method with two configurations of the Qwen-based projector: (1) without attribute guidance, only use categorical emotion supervision, (2) without any training. The result is shown in~\Cref{tab:model_com} and \Cref{fig:illustrate}. The attribute-aware configuration achieved the highest scores across all metrics. Correspondingly, we visualize the DES embeddings using MDS projection, as shown in~\Cref{fig:cloud}. DES embeddings trained with full attribution form the most compact and clear clusters, showing that including attribute information helps the DES space to encode more detailed emotional semantics and contextual dependencies.

\section{Conclusion}
\label{sec:conclusion}

In this work, we introduced EmoVerse, a large-scale and interpretable visual emotion dataset designed to advance fine-grained affective understanding. By integrating diverse sources and constructing multi-level annotations, EmoVerse offers a solid foundation for interpretable visual analysis. The multi-stage verification pipeline ensures the accuracy of our annotations. Building on these annotations, we further developed an interpretable emotion projector that maps visual cues into a high-dimensional DES space and provides interpretable explanations for emotion understanding.

\begin{sloppypar}
In future works, We plan to extend EmoVerse to multi-emotion scenarios, integrate multimodal cues, and enable emotion-controllable generation. We hope EmoVerse can serve as a strong benchmark and inspire future research on interpretable Visual Emotion Analysis.
\end{sloppypar}
{
    \small
    \bibliographystyle{ieeenat_fullname}
    \bibliography{reference}
}

\end{CJK*}
\end{document}